\documentclass[lettersize,journal]{IEEEtran}
\usepackage{amsmath,amsfonts}
\usepackage{algorithmic}
\usepackage{array}
\usepackage[caption=false,font=normalsize,labelfont=sf,textfont=sf]{subfig}
\usepackage{textcomp}
\usepackage{stfloats}
\usepackage{url}
\usepackage{verbatim}
\usepackage{graphicx}

\def\BibTeX{{\rm B\kern-.05em{\sc i\kern-.025em b}\kern-.08em
    T\kern-.1667em\lower.7ex\hbox{E}\kern-.125emX}}
\usepackage{times}
\usepackage{multicol}

\usepackage{cite}
\usepackage{microtype}
\usepackage{amsmath,amssymb,amsfonts}
\usepackage{algorithmic}
\usepackage{hhline}
\usepackage{subcaption}
\usepackage{itmp}
\usepackage{sequat}
\usepackage{dyatex}

\usepackage{float}
\usepackage{boldline,multirow}
\usepackage{siunitx}

\usepackage[caption=false,font=normalsize,labelfont=sf,textfont=sf]{subfig}
\usepackage{mwe}
\usepackage{tikz}
\usetikzlibrary{decorations.pathreplacing}
\usetikzlibrary{decorations.markings}
\usepackage[percent]{overpic}
\usepackage[algoruled,vlined,linesnumbered]{algorithm2e}
\usepackage{xcolor}
\usepackage{ulem}
\usepackage{tkz-euclide}
\usepackage{minescolors}
\usepackage{pgfplots}
\usepackage{balance}

\makeatletter
\let\NAT@parse\undefined
\makeatother
\usepackage[
bookmarks=true,
pdfusetitle,
colorlinks,
bookmarksopen,
bookmarksnumbered,
citecolor=blue,
urlcolor=blue,
linkcolor=blue
]{hyperref}

\newcommand{\Qseeds}{Q_{\mathrm{seeds}}}

\renewcommand{\conf}[0]{\ensuremath{\boldsymbol{q}}}
\newcommand{\qstart}[0]{\ensuremath{\conf_{\mathrm{start}}}}
\newcommand{\qgoal}[0]{\ensuremath{\conf_{\mathrm{goal}}}}
\newcommand{\CSpace}[0]{\ensuremath{\mathcal{C}}}
\newcommand{\CObs}[0]{\ensuremath{\mathcal{C}_{\mathrm{obs}}}}
\newcommand{\CFree}[0]{\ensuremath{\mathcal{C}_{\mathrm{free}}}}

\newcommand{\infproof}[0]{\ensuremath{\mathcal{M}}}

\renewcommand{\Re}[0]{\mathbb{R}}
\renewcommand{\emph}[1]{{\textit{#1}}}
\newcommand{\Tri}[0]{\mathbb{CTR}}

\SetKwComment{Pragma}{\#pragma }{}

\newcommand{\tabtm}[1]{\itshape{#1}}
\newcommand{\tablbl}[1]{\itshape{#1}}

\SetKwProg{Fn}{function}{\ is}{end}
\SetKwBlock{Loop}{loop}{end}

\SetCommentSty{mycommfont}
\SetKwSty{mykwfont}

\dyavenueshort
\def\BibTeX{{\rm B\kern-.05em{\sc i\kern-.025em b}\kern-.08em
    T\kern-.1667em\lower.7ex\hbox{E}\kern-.125emX}}
\usepackage[top=60pt,bottom=43pt,left=48pt,right=48pt]{geometry}

\usepackage{balance}

\begin{document}
\title{Scaling Motion Planning Infeasibility Proofs}
\author{
  Sihui Li$^{1}$ \and and Neil T. Dantam$^{1}$%
  \thanks{
This work was supported by NSF IIS-1849348, %
    NSF CCF-2124010, %
    ONR N00014-21-1-2418, and ARL TBAM-CRP [W911NF-
22-2-0235].}
\thanks{$^{1}$ Sihui Li and Neil T. Dantam are with the Department of Computer Science,
    Colorado School of Mines, USA.
    {\tt\footnotesize {\{li, ndantam\}@mines.edu}}}%
}


\maketitle






\begin{abstract}
Achieving completeness in the motion planning problem demands substantial computation power, especially in high dimensions. Recent developments in parallel computing have rendered this more achievable. We introduce an embarrassingly parallel algorithm for constructing infeasibility proofs. Specifically, we design and implement a manifold triangulation algorithm on GPUs based on manifold tracing with Coxeter triangulation. To address the challenge of extensive memory usage within limited GPU memory resources during triangulation, we introduce batch triangulation as part of our design. The algorithm provides two orders of magnitude speed-up compared to the previous method for constructing infeasibility proofs. The resulting asymptotically complete motion planning algorithm effectively leverages the computational capabilities of both CPU and GPU architectures and maintains minimum data transfer between the two parts. We perform experiments on 5-DoF and 6-Dof manipulator scenes. 
\end{abstract}


\section{Introduction}
A complete motion planer returns a plan or report plan non-existence in finite time. Previous work has shown that motion planning is PSPACE-hard~\cite{reif1979complexity}, which means solving the motion planning problem completely is an inherently computationally heavy process in high dimensions. Because of this, most previous works in motion planning~\cite{lavalle1998rapidly,karaman2011sampling,kavraki1996probabilistic,kuffner2000rrt,sucan2012open, shkolnik2011sample} focus on solving one part of this problem - finding a valid plan. The second part, which is proving path non-existence, is largely omitted. 
Path non-existence guarantees, or what we call infeasibility proofs, are becoming more crucial as we move into higher-level planning problems~\cite{ben1998practical,cambon2009hybrid,ota2004rearrangement,wilfong1991motion}, where motion planning is a sub-routine.

In our previous work, we proposed an asymptotically complete algorithm framework~\cite{li2022exponential, li2023framework, li2023scaling}. A motion planner is asymptotically complete if it returns a plan or an infeasibility proof given long enough time. The algorithm runs alone-side a sampling-based motion planner~\cite{kavraki1996probabilistic}, finds a plan if a plan exists, or returns an infeasibility proof otherwise. For finding infeasibility proofs, the core idea is to create a ``closure'' in the obstacle region of the configuration space that separates the start and the goal. The algorithm first learns a manifold from samples in the configuration space, and then triangulation the manifold to make sure it is entirely in the obstacle region. The algorithm scales up to 5-DoF~\cite{li2023scaling}. The main limitation is on the manifold triangulation step and collision checking, which takes over $95\%$ of the total runtime. 

To deal with the increased computation needs in higher dimensions, one way is to take advantage of multi-core or specialized hardware. Prior work has used
multi-core CPUs to parallelize
sampling~\cite{ichnowski2012parallel,plaku2005sampling} or nearest
neighbor search~\cite{ichnowski2020concurrent}, GPUs for nearest
neighbor search~\cite{pan2010efficient} and collision
checking~\cite{pan2012gpu}, and FPGAs for collision
checking~\cite{murray2016robot}.

\usetikzlibrary{shapes.arrows}
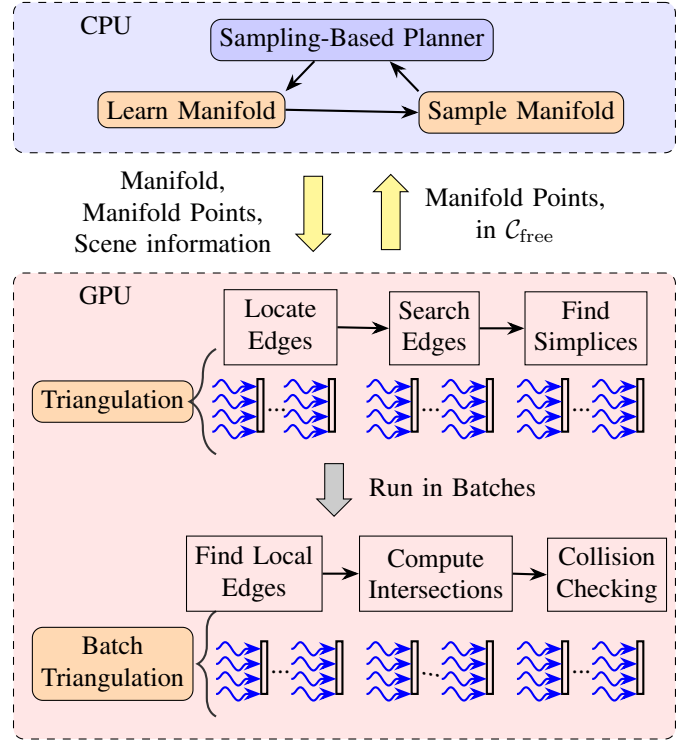
\begin{figure}
    \centering
    \begin{tikzpicture}
    [node_style/.style={draw, rectangle, rounded corners},
    inf_node/.style={node_style, fill=orange!30},
    arrow_style/.style={thick, -Stealth}]
    \node[node_style,fill=blue!10, dashed, minimum width = 8.8cm, minimum height=2cm] (cpu) at (-0.1cm, -0.5cm) {};
        \node[node_style,fill=blue!20] (planner) at (0, 0) {Sampling-Based Planner};
        \coordinate [label={left:CPU}] (s0) at (-2.8cm, 0.2cm);
        \node[inf_node, 
        below left=0.4cm and -1cm of planner] (learnm) {Learn Manifold};
        \node[inf_node, below right=0.4cm and -1cm of planner] (samplem) {Sample Manifold};
        \draw[->, arrow_style] ($(planner.south) - (0.5cm, 0cm)$) -- (learnm.north east);
        \draw[->, arrow_style] (samplem.north west)--($(planner.south) + (0.5cm, 0cm)$);
        \draw[arrow_style] (learnm.east) -- (samplem.west);
        
\node[single arrow, draw=black, fill=yellow!50, 
      minimum width = 8pt, single arrow head extend=3pt,
      minimum height=10mm, below left=2.1cm and -1.5cm of planner, rotate=-90] (cpu2gpu) {};
\node[single arrow, draw=black, fill=yellow!50, 
      minimum width = 8pt, single arrow head extend=3pt,
      minimum height=10mm, below right=2.2cm and -1.5cm of planner, rotate=90] (gpu2cpu){};
\node[align=center, below left=-0.5 cm and 0.3cm of cpu2gpu, minimum width=1.5cm, minimum height=1cm] {Manifold, \\Manifold Points, \\Scene information};
\node[align=center, below right=-0.4cm and 0.2cm of gpu2cpu, minimum width=1.5cm, minimum height=1cm] {Manifold Points, \\in $\CFree$};
\node[node_style,fill=red!10, dashed, minimum width = 8.8cm, minimum height=6.2cm] (gpu) at (-0.1cm, -6.2cm) {};
\coordinate [label={left:GPU}] (s0) at (-2.8cm, -3.4cm);
        \node[inf_node, above left =-2cm and -2.4cm of gpu] (triangulation){Triangulation};
        \draw [black!80,thick,
    decorate, 
    decoration = {brace,
        amplitude=10pt}] ($(triangulation.east) + (0.3cm, -0.7cm)$) --  ($(triangulation.east) + (0.3cm, 0.7cm)$);

        \begin{scope}[
    	xshift=0, every node/.append style={
    	    yslant=0.5,xslant=-1},yslant=0,xslant=0
    	  ]
       \def\x{1.2}
       \def\y{-0.4}
       \def\dx{0.6}
       \def\dy{0.7}
    \draw[black, thick] ($(triangulation.east) + (\x + \dx, \y)$) rectangle  ($(triangulation.east) + (\x+\dx + 0.08, \y+\dy)$);
\foreach \i in {0,...,3}
        {\path[draw=blue, very thick, decorate, 
    decoration = {snake,   
                    pre length=0.3pt,post length=5pt,
                    }, arrow_style] ($(triangulation.east) + (\x, \y + \i * 0.2)$) --  ($(triangulation.east) + (\x + \dx, \y + \i * 0.2)$);}
    
    \end{scope}
        \begin{scope}[
    	xshift=0, every node/.append style={
    	    yslant=0,xslant=0},yslant=0,xslant=0
    	  ]
       \def\x{0.25}
       \def\y{-0.4}
       \def\dx{0.6}
       \def\dy{0.7}
        \draw[black, thick] ($(triangulation.east) + (\x + \dx, \y)$) rectangle  ($(triangulation.east) + (\x+\dx + 0.08, \y+\dy)$);
        \foreach \i in {0,...,3}
        {\path[draw=blue, very thick, decorate, 
    decoration = {snake,   
                    pre length=0.3pt,post length=5pt,
                    }, arrow_style] ($(triangulation.east) + (\x, \y + \i * 0.2)$) --  ($(triangulation.east) + (\x + \dx, \y + \i * 0.2)$);}
        \node[align=center, above right=\y+0.15 and \x+\dx of triangulation.east] {...};
        
    \end{scope}

\node[draw, align=center, above right=0.2cm and 0.4cm of triangulation, minimum width=1.5cm, minimum height=1cm] (le){Locate \\Edges};

        \begin{scope}[
    	xshift=0, every node/.append style={
    	    yslant=0,xslant=0},yslant=0,xslant=0
    	  ]
       \def\x{3.3}
       \def\y{-0.4}
       \def\dx{0.6}
       \def\dy{0.7}
            \draw[black, thick] ($(triangulation.east) + (\x + \dx, \y)$) rectangle  ($(triangulation.east) + (\x+\dx + 0.08, \y+\dy)$);
\foreach \i in {0,...,3}
        {\path[draw=blue, very thick, decorate, 
    decoration = {snake,   
                    pre length=0.3pt,post length=5pt,
                    }, arrow_style] ($(triangulation.east) + (\x, \y + \i * 0.2)$) --  ($(triangulation.east) + (\x + \dx, \y + \i * 0.2)$);}
    \end{scope}
    
        \begin{scope}[
    	xshift=0, every node/.append style={
    	    yslant=0,xslant=0},yslant=0,xslant=0
    	  ]
       \def\x{2.3}
       \def\y{-0.4}
       \def\dx{0.6}
       \def\dy{0.7}
\draw[black, thick] ($(triangulation.east) + (\x + \dx, \y)$) rectangle  ($(triangulation.east) + (\x+\dx + 0.08, \y+\dy)$);
        \foreach \i in {0,...,3}
        {\path[draw=blue, very thick, decorate, 
    decoration = {snake,   
                    pre length=0.3pt,post length=5pt,
                    }, arrow_style] ($(triangulation.east) + (\x, \y + \i * 0.2)$) --  ($(triangulation.east) + (\x + \dx, \y + \i * 0.2)$);}
        \node[align=center, above right=\y+0.15 and \x+\dx of triangulation.east] {...};
    \end{scope}
\node[draw, align=center, above right=0.2cm and 2.6cm of triangulation] (se) {Search \\Edges};

        \begin{scope}[
    	xshift=0, every node/.append style={
    	    yslant=0.5,xslant=-1},yslant=0,xslant=0
    	  ]
       \def\x{5.3}
       \def\y{-0.4}
       \def\dx{0.6}
       \def\dy{0.7}
    \draw[black, thick] ($(triangulation.east) + (\x + \dx, \y)$) rectangle  ($(triangulation.east) + (\x+\dx + 0.08, \y+\dy)$);
\foreach \i in {0,...,3}
        {\path[draw=blue, very thick, decorate, 
    decoration = {snake,   
                    pre length=0.3pt,post length=5pt,
                    }, arrow_style] ($(triangulation.east) + (\x, \y + \i * 0.2)$) --  ($(triangulation.east) + (\x + \dx, \y + \i * 0.2)$);}
    \end{scope}
        \begin{scope}[
    	xshift=0, every node/.append style={
    	    yslant=0,xslant=0},yslant=0,xslant=0
    	  ]
       \def\x{4.3}
       \def\y{-0.4}
       \def\dx{0.6}
       \def\dy{0.7}
\draw[black, thick] ($(triangulation.east) + (\x + \dx, \y)$) rectangle  ($(triangulation.east) + (\x+\dx + 0.08, \y+\dy)$);
        \foreach \i in {0,...,3}
        {\path[draw=blue, very thick, decorate, 
    decoration = {snake,   
                    pre length=0.3pt,post length=5pt,
                    }, arrow_style] ($(triangulation.east) + (\x, \y + \i * 0.2)$) --  ($(triangulation.east) + (\x + \dx, \y + \i * 0.2)$);}
        \node[align=center, above right=\y+0.15 and \x+\dx of triangulation.east] {...};
    \end{scope}
    
\node[draw, align=center, above right=0.2cm and 4.4cm of triangulation] (fs) {Find \\Simplices};

\node[single arrow, draw=black, fill=black!20, 
      minimum width = 8pt, single arrow head extend=3pt,
      minimum height=7mm, below right=0.7cm and 2cm of triangulation, rotate=-90] {}; 


\node[inf_node, minimum width= 2cm, align=center, above left =-5.7cm and -2.4cm of gpu] (batchtri) {Batch \\Triangulation};
        \draw [black!80,thick,
    decorate, 
    decoration = {brace,
        amplitude=7pt}] ($(batchtri.east) + (0.3cm, -0.7cm)$) --  ($(batchtri.east) + (0.3cm, 0.7cm)$);

        \begin{scope}[
    	xshift=0, every node/.append style={
    	    yslant=0,xslant=-1},yslant=0,xslant=0
    	  ]
       \def\x{1.3}
       \def\y{-0.4}
       \def\dx{0.6}
       \def\dy{0.7}
            \draw[black, thick] ($(batchtri.east) + (\x + \dx, \y)$) rectangle  ($(batchtri.east) + (\x+\dx + 0.08, \y+\dy)$);
\foreach \i in {0,...,3}
        {\path[draw=blue, very thick, decorate, 
    decoration = {snake,   
                    pre length=0.3pt,post length=5pt,
                    }, arrow_style] ($(batchtri.east) + (\x, \y + \i * 0.2)$) --  ($(batchtri.east) + (\x + \dx, \y + \i * 0.2)$);}
    \end{scope}
        \begin{scope}[
    	xshift=0, every node/.append style={
    	    yslant=0,xslant=0},yslant=-0,xslant=0
    	  ]
       \def\x{0.3}
       \def\y{-0.4}
       \def\dx{0.6}
       \def\dy{0.7}
\draw[black, thick] ($(batchtri.east) + (\x + \dx, \y)$) rectangle  ($(batchtri.east) + (\x+\dx + 0.08, \y+\dy)$);
        \foreach \i in {0,...,3}
        {\path[draw=blue, very thick, decorate, 
    decoration = {snake,   
                    pre length=0.3pt,post length=5pt,
                    }, arrow_style] ($(batchtri.east) + (\x, \y + \i * 0.2)$) --  ($(batchtri.east) + (\x + \dx, \y + \i * 0.2)$);}
        \node[align=center, above right=\y+0.15 and \x+\dx of batchtri.east] {...};
    \end{scope}

\node[draw, align=center, above right=0.2cm and -0.1cm of batchtri, minimum width=1.5cm, minimum height=1cm] (fle) {Find Local \\Edges};

        \begin{scope}[
    	xshift=0, every node/.append style={
    	    yslant=0,xslant=0},yslant=0,xslant=0
    	  ]
       \def\x{3.3}
       \def\y{-0.4}
       \def\dx{0.6}
       \def\dy{0.7}
    \draw[black, thick] ($(batchtri.east) + (\x + \dx, \y)$) rectangle  ($(batchtri.east) + (\x+\dx + 0.08, \y+\dy)$);
\foreach \i in {0,...,3}
        {\path[draw=blue, very thick, decorate, 
    decoration = {snake,   
                    pre length=0.3pt,post length=5pt,
                    }, arrow_style] ($(batchtri.east) + (\x, \y + \i * 0.2)$) --  ($(batchtri.east) + (\x + \dx, \y + \i * 0.2)$);}
    \end{scope}
        \begin{scope}[
    	xshift=0, every node/.append style={
    	    yslant=-0.15,xslant=0},yslant=0,xslant=0
    	  ]
       \def\x{2.3}
       \def\y{-0.4}
       \def\dx{0.6}
       \def\dy{0.7}
\draw[black, thick] ($(batchtri.east) + (\x + \dx, \y)$) rectangle  ($(batchtri.east) + (\x+\dx + 0.08, \y+\dy)$);
        \foreach \i in {0,...,3}
        {\path[draw=blue, very thick, decorate, 
    decoration = {snake,   
                    pre length=0.3pt,post length=5pt,
                    }, arrow_style] ($(batchtri.east) + (\x, \y + \i * 0.2)$) --  ($(batchtri.east) + (\x + \dx, \y + \i * 0.2)$);}
        \node[align=center, above right=\y+0.15 and \x+\dx of batchtri.east] {...};
    \end{scope}
\node[draw, align=center, above right=0.2cm and 2.2cm of batchtri, minimum height=1cm] (ci) {Compute \\Intersections};

        \begin{scope}[
    	xshift=0, every node/.append style={
    	    yslant=0,xslant=0},yslant=0,xslant=0
    	  ]
       \def\x{5.3}
       \def\y{-0.4}
       \def\dx{0.6}
       \def\dy{0.7}
    \draw[black, thick] ($(batchtri.east) + (\x + \dx, \y)$) rectangle  ($(batchtri.east) + (\x+\dx + 0.08, \y+\dy)$);
\foreach \i in {0,...,3}
        {\path[draw=blue, very thick, decorate, 
    decoration = {snake,   
                    pre length=0.3pt,post length=5pt,
                    }, arrow_style] ($(batchtri.east) + (\x, \y + \i * 0.2)$) --  ($(batchtri.east) + (\x + \dx, \y + \i * 0.2)$);}
    \end{scope}
        \begin{scope}[
    	xshift=0, every node/.append style={
    	    yslant=0,xslant=0},yslant=0,xslant=0
    	  ]
       \def\x{4.3}
       \def\y{-0.4}
       \def\dx{0.6}
       \def\dy{0.7}
\draw[black, thick] ($(batchtri.east) + (\x + \dx, \y)$) rectangle  ($(batchtri.east) + (\x+\dx + 0.08, \y+\dy)$);
        \foreach \i in {0,...,3}
        {\path[draw=blue, very thick, decorate, 
    decoration = {snake,   
                    pre length=0.3pt,post length=5pt,
                    }, arrow_style] ($(batchtri.east) + (\x, \y + \i * 0.2)$) --  ($(batchtri.east) + (\x + \dx, \y + \i * 0.2)$);}
        \node[align=center, above right=\y+0.15 and \x+\dx of batchtri.east] {...};
    \end{scope}
\node[draw, align=center, above right=0.2cm and 4.7cm of batchtri] (cc) {Collision \\Checking};
\draw[->, arrow_style] (le.east)--(se.west);
\draw[->, arrow_style] (se.east)--(fs.west);
\draw[->, arrow_style] (fle.east)--(ci.west);
\draw[->, arrow_style] (ci.east)--(cc.west);

\coordinate [label={above: Run in Batches}] (rb) at ($(ci.east) + (-0.8, 0.9)$);

    \end{tikzpicture}

    \caption{Algorithm overview. The algorithm maximizes resource utilization by efficiently distributing its components across both CPU and GPU. The triangulation and the collision checking steps are completed entirely on GPU. }
    \label{fig:diagram}
\end{figure}

\dyathesis{In this work, we present an embarrassingly parallel algorithm for constructing infeasibility proofs on GPUs.} 
This algorithm runs two orders of magnitudes faster than our previous algorithm through vast parallelism in the triangulation step and collision checking. 
\autoref{fig:diagram} shows the overall algorithm structure. The entire triangulation and collision checking step runs on GPU, which ensures minimum data transfer between CPU and GPU. 
We design the triangulation algorithm based on manifold tracing with Coxeter triangulation~\cite{boissonnat2023tracing} to fit the computation architecture of GPUs. To handle the significant memory requirements within the constraints of limited GPU memory during triangulation, we design the algorithm to support batch triangulation. We also perform collision checking on GPU based on spherical robots and primary-shaped environments. We perform experiments on up to 6-DoF robots to show the scalability of our algorithm.  

\section{Related Work}

\subsection{Infeasibility Proof}
Many sampling-based motion planners are probabilistically complete~\cite{lavalle1998rapidly,karaman2011sampling,kavraki1996probabilistic,kuffner2000rrt,sucan2012open,
  shkolnik2011sample}, meaning they can find a plan given long enough time. However, they cannot provide path non-existence guarantees. There are previous works that construct exact path non-existence
guarantees. {\cite{varava2020caging}} proves path non-existence for
single, rigid objects in a 2D or 3D workspace to guarantee stable
grasp.
{\cite{basch2001disconnection}} considers the specific problem of a
rigid body passing through a narrow gate.
{\cite{mccarthy2012proving}} constructs alpha-shapes in the obstacle
region to query the connectivity of two configurations, which works
for up to 3-dimensional configuration spaces. These methods do not
apply to general manipulators' configuration spaces.

There are also methods that provide approximate path non-existence guarantees.
In visibility~\cite{simeon2000visibility} and
sparsity~\cite{dobson2014sparse} based planners, 
if no plan is found when the algorithm terminates, the problem
may be considered infeasible to some extent since they achieve high coverage of the free space~\cite{orthey2021sparse}.
In deterministic sampling-based motion planning, if no plan is found, then either
no solution exists or a solution exists only through some narrow
passages~\cite{branicky2001quasi,janson2018deterministic,tsao2020sample, dayan2023near}. 
Previous works have also proposed learning-based methods to
predict infeasible
plans~\cite{wells2019learning,dreiss2020deep,driess2021learninggeometric,driess2021learningsequential,bouhsain2023learning}.
All of these methods do not provide exact plan non-existence guarantees.

Previously, we have done a series of work on constructing infeasibility proofs in general configuration spaces. In~\cite{li2020towards}, we defined infeasibility proof and proposed our first method for infeasibility proof construction. Based on this work, we proposed a learning and sampling framework to construct infeasibility proof~\cite{li2021learning}. We scaled this work by adapting Coxeter triangulation~\cite{kachanovich2019meshing} into our framework~\cite{li2023scaling}, which eventually got the algorithm to run for up to 5-DoF manipulators' configuration space. In this work, we scale this algorithm framework further to run in 6-D by re-designing the triangulation algorithm for GPU parallelization. 

\subsection{Parallel Computing in Motion Planning}
Because motion planning is a computational-heavy problem, previous work has made various efforts in paralleling the processes to improve runtime. In this section, we provide an overview of parallelism utilization in prior works, categorizing them according to hardware types, as parallelizing schemes could differ significantly across different hardware platforms. 

Parallelism on CPUs usually happens at the algorithmic level, where tasks are divided into smaller, independent units that can be executed simultaneously. Early works~\cite{challou1993parallel, challou1995parallel} looked at how to distribute search efforts among many processors based on gradient descent and random walk. 
Previous work has also looked at parallelizing sampling-based motion planners~\cite{ichnowski2012parallel, amato1999probabilistic, plaku2005sampling, jacobs2012scalable, devaurs2011parallelizing, jacobs2013scalable}, through parallelizing sampling and growth of the search tree or graph; and parallelizing search of decomposed configuration space. Search-based planning has also seen application with parallel computing~\cite{phillips2014pa, mukherjee2022mplp}. CPU fine-grained parallelization is also possible. In a recent work~\cite{thomason2023motions}, the authors achieve microsecond-level motion planning through fine-grained parallelism for forward kinematics and collision checking using CPU SIMD instructions. 

Modern GPUs have a significantly higher number of cores compared to CPUs, enabling the potential for greater speed-ups. Recent work~\cite{sundaralingam2023curobo} provides a GPU-accelerated library that runs trajectory optimization in parallel with many seeds. Previous work has explored the utilization of GPUs to accelerate particular time composing parts of motion planning such as collision checking and nearest neighbor calculation~\cite{pan2012gpu, pan2010efficient}. Computation of Generalized Voronoi Diagrams on GPUs has also been explored for real-time motion planning~\cite{hoff2000interactive, sud2008real}.  
Field-programmable gate arrays (FPGAs) are employed in motion planing~\cite{murray2016robot, murray2016microarchitecture}, substantially speeding up collision checking and edge validation. 
In this work, we employ GPUs to construct infeasibility proofs. 

\section{Problem Definition}
\label{sec:problem}
The inputs of motion planning
problem~\cite{lavalle2006planning} consist of a configuration space
$\CSpace$ of dimension $n$, a start configuration $\qstart$, and a
goal configuration $\qgoal$.  The configuration space $\CSpace$ is the
union of the disjoint obstacle region $\CObs$ and free space
$\CFree$. 
$\CObs$ and $\CFree$ are
implicitly defined through a validity checking function. Given a configuration, the function returns whether the robot in the configuration collides
with obstacles or not.  Both $\qstart$ and
$\qgoal$ are in $\CFree$.
The output is either a plan or an infeasibility proof in the limit.  
When a plan exists, the output is a plan $\sigma$ such that
$\sigma[0, 1] \in \CFree$, $\sigma[0] = \qstart$,
$\sigma[1] = \qgoal$.  When there is no feasible plan, the output is
an infeasibility proof.  An infeasibility proof
is a closed manifold defined through an implicit function $\infproof(\conf) = 0$ that lies entirely in the obstacle
region, $\{\conf \mid \infproof(\conf) = 0\} \subseteq \mathcal{C}_{obs}\}$, and separates $\qstart$ and $\qgoal$, $\infproof(\qstart) \infproof(\qgoal) < 0$~\cite{li2023framework}.

In this work, we consider kinematic problems only, that is, $\CObs$ is only caused by robot and obstacle geometries in the workspace. Also, we assume an Euclidean configuration space instead of a
\emph{metric space} (e.g. $\SEthree$). More details on configuration space requirements are in~\cite{li2022exponential}.

\section{Background}
\subsection{Infeasibility Proof Construction}
\label{sec:inf}
This work is based on a previous algorithm framework we proposed~\cite{li2023framework, li2023scaling} that combines learning and sampling to construct infeasibility proofs. The algorithm runs in parallel with sampling-based motion
planners (i.e. PRM). We treat the sampled configurations as data. One class contains all samples connectable to the start (or goal); The other class contains all other samples. Then we train a classifier with these two classes of samples. The resulting manifold converges to an infeasibility proof as we defined in \autoref{sec:problem}~\cite{li2022exponential}. Then we triangulate the manifold to produce a piece-wise linear approximation of the manifold. Lastly, we check that the triangulation is entirely in $\CObs$ with configuration space penetration depth~\cite{li2021learning} or simplex decomposition~\cite{li2023scaling}. 
The major scalability limit of this algorithm is the demanding computation time for the triangulation and checking step in high dimensions. In this work, we address this challenge by employing GPU computation for both triangulation and collision checking.

\subsection{Permutahedral Representation}
\label{sec:pr}
This section provides a brief introduction of the permutahedral representation~\cite{boissonnat2023tracing} used in the manifold tracing algorithm with Coxeter triangulation. 
Coxeter triangulation is a way to triangulate any $n$-dimensional Euclidean space $\Re^n$, and is defined using root systems~\cite{kachanovich2019meshing}. A root system can be viewed as a set of basis vectors for the triangulation. Permutahedral representation~\cite{boissonnat2023tracing} provides a compact way to represent simplices in $\Re^n$ Coxeter triangulation. 
For a $k$-simplex in $\Re^n$ Coxeter triangulation, the representation has two parts, a vertex and a partition. 
The vertex is the smallest vertex of the $k$-simplex in the lexicographical order, represented with a list of $n$ integer values $[v_0, v_1, ..., v_{n-1}]$ that corresponds to the coordinate of the vertex in $\Re^n$. 
The partition is an ordered partition of $\{0, 1, ..., n\}$ in $k+1$ parts, and represents the cycle of how other vertices of the simplex are located using the triangulation's root system. \autoref{fig:pr} shows the permutahedral representation of two $2$-simplices in a square. 
The permutahedral representation supports several essential queries within the triangulation algorithm based on manifold tracing. These queries include finding vertices, faces, and cofaces of a simplex, as well as locating points within $n$-simplices. For further details on how to perform these queries, please refer to~\cite{boissonnat2023tracing}.

\newcommand{\pythagwidth}{2cm}
\newcommand{\pythagheight}{2cm}
\tikzset{
  on each segment/.style={
    decorate,
    decoration={
      show path construction,
      moveto code={},
      lineto code={
        \path [#1]
        (\tikzinputsegmentfirst) -- (\tikzinputsegmentlast);
      },
      curveto code={
        \path [#1] (\tikzinputsegmentfirst)
        .. controls
        (\tikzinputsegmentsupporta) and (\tikzinputsegmentsupportb)
        ..
        (\tikzinputsegmentlast);
      },
      closepath code={
        \path [#1]
        (\tikzinputsegmentfirst) -- (\tikzinputsegmentlast);
      },
    },
  },
  mid arrow/.style={postaction={decorate,decoration={
        markings,
        mark=at position .5 with {\arrow[#1]{stealth}}
      }}},
  left arrow/.style={postaction={decorate,decoration={
        markings,
        mark=at position .7 with {\arrow[#1]{stealth}}
      }}},
}
\begin{figure}
    \centering
    \begin{tikzpicture}
\coordinate [label={below left:$v_0: [0, 0]$}] (v0) at (0, 0);
\coordinate [label={below right:$v_1$: [1, 0]}] (v1) at (\pythagwidth, 0);
\coordinate [label={above right:$v_2$: [1, 1]}] (v2) at (\pythagwidth, \pythagheight);
\coordinate [label={above left:$v_3$: [0, 1]}] (v3) at (0, \pythagheight);
\coordinate [label={above left:$v: [0, 0]$}] (s0) at (\pythagwidth + 1.45cm, \pythagheight/2);
\coordinate [label={above left:$p:\{\{0\}, \{1\}, \{2\}\}$}] (s0) at (\pythagwidth + 3cm, \pythagheight/2-0.4cm);
\coordinate [label={above left:$v: [0, 0]$}] (s1) at (\pythagwidth - 3.5cm, \pythagheight/2);
\coordinate [label={above left:$p:\{\{1\}, \{0\}, \{2\}\}$}] (s1) at (\pythagwidth - 2cm, \pythagheight/2-0.4cm);

\path[draw, fill=red!30!white, line width=1pt, postaction={on each segment={mid arrow=black}}]  (v0) -- (v1)--(v2)--cycle;  
\path[draw, fill=blue!30!white, line width=1pt, postaction={on each segment={mid arrow=black}}]  (v0) -- (v3)--(v2)--cycle;  
\end{tikzpicture}  
    \caption{Example 2D permutahedral representation, ``v'' for vertex, ``p'' for partition, for the two triangles (blue and red) in a square. }
    \label{fig:pr}
\end{figure}
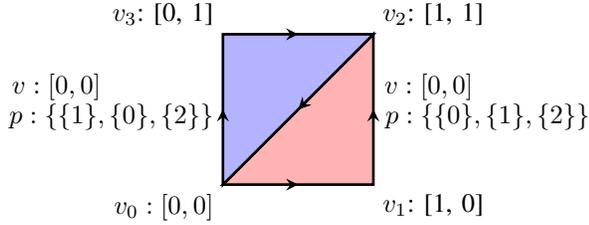

\section{Algorithm}
The algorithm leverages the computational power of both CPUs and GPUs. We keep the parts that already run reasonably fast and/or require a lot of branching on the CPU, and move the highly parallelizable and time-consuming parts to the GPU. \autoref{fig:diagram}
shows the overall algorithm structure. 
Using this structure, we can maintain minimum data transfer between the CPU and the GPU. We run manifold learning and manifold sampling on CPU. The CPU transfers the learned manifold and a set of manifold points to the GPU for triangulation, and robot kinematic and world geometry information to the GPU for collision checking. The triangulation and collision checking steps are completed on GPU. The computation on the GPU generates a large amount of data but only needs to transfer the $\CFree$ manifold points to the CPU. 
Previously, we also need to decompose the facets of the triangulation for collision checking within a minimum distance. In this work, batch triangulation in the triangulation algorithm removes the need for facet decomposition, as batch triangulation can produce a sufficiently refined triangulation.

\subsection{Triangulation on GPU}
\label{sec:trigpu}
In this section, we discuss the GPU triangulation algorithm in detail. Contemporary GPUs offer greater parallel processing power compared to CPUs. This makes GPUs particularly well-suited for manifold triangulation tasks, which have a lot of parallelizable components. 
On the other hand, GPUs have specialized architecture and memory hierarchy compared to CPUs. For GPUs to perform optimally, we need to consider these characteristics when designing our algorithm. 

First, GPU threads are SIMD (single-instruction multiple-data). It works best when all threads perform the same instructions on different pieces of data with as less diverging branches as possible. To take advantage of this architecture, it is best to run a large number of threads at one time that each performs simple tasks so that all GPU cores are fully utilized. 
Secondly, GPUs usually have very limited shared memory resources between threads. This requires us to carefully manage memory to ensure that the memory consumption remains within limits while all threads still have the necessary resources to execute the tasks. This is especially important for the triangulation step since some of the intermediate steps in triangulation require a large amount of memory. We discuss this further in \autoref{sec:batchtri}. 

Our GPU triangulation algorithm is based on the manifold tracing algorithm with Coxeter triangulation~\cite{boissonnat2023tracing, boissonnat2022topological}. 
The manifold tracing algorithm produces a piece-wise linear approximation of a $m$-dimensional smooth manifold in $\Re^n$. The inputs to the manifold tracing algorithm are the
manifold, in our case the learned manifold as described in \autoref{sec:inf}, and a set of points on the manifold. 
These points are the seeds for constructing the triangulation, noted by $\Qseeds$. We obtain $\Qseeds$ by sampling points on the manifold, which involves solving a nonlinear optimization program that minimizes the absolute value of the manifold function with a given configuration. 
The output of the manifold tracing algorithm is a set of $k$-dimensional simplices that intersect the manifold, where $k=n-m$ is called the codimension. In our case, the learned manifold is always
($n$-$1$)-dimensional for $\Re^n$ configuration space. This means the
codimension is always $1$ and the output simplices are always
$1$-simplices, i.e., line segments. 

The algorithm starts with a Coxeter triangulation of the entire $\Re^n$ space ($\Tri$ in the following). 
First, the algorithm locates the
$n$-simplices containing $\Qseeds$ by querying the permutahedral
representation of $\Tri$.  Then, the algorithm saves all $k$-simplices that are
in the located $n$-simplices and intersect the manifold. 
This process requires face querying using permutahedral representation. 
Starting from these $k$-simplices, the algorithm runs
breadth-first search on $\Tri$ by visiting neighboring $k$-simplices
of the saved $k$-simplices and saving all neighbors that intersect the
manifold.  A $k$-simplex is considered a neighbor of another
$k$-simplex if they share the same ($k$+$1$)-simplices. Finding neighboring $k$-simplex requires coface and face querying. 
The algorithm terminates when graph traversal ends, and outputs the $k$-simplices and their intersection points with the manifold. A complete
description is
in~\cite{kachanovich2019meshing}.
$\Re^n$ triangulation size is adjustable with a parameter $\lambda_T$,
which also determines the final manifold triangulation size. Smaller
triangles produce intersection points that are closer to each other, \ie{},
a finer manifold triangulation.

To make this algorithm run on GPU, the first step is to perform the same types of queries for vertices, faces, cofaces, and locating points. 
On CPUs, these queries are the same functions that work for any dimension, which requires dynamic memory allocations. 
On GPUs, we want to avoid dynamic memory management since it is expansive, can lead to fragmentation, and requires changes to the manageable memory size limit~\cite{winter2021dynamic}. 
Since our co-dimension is always $1$, we only need limited types of queries and simplices, which means we can design fixed-size versions of these queries and allocate fixed-size shared memory for the computation. 
The types of queries we need are, 1) locating $n$-simplices for points; 2) querying edges ($1$-simplex faces) of $n$-simplices; 3) querying $2$-simplex cofaces of $1$-simplices; 4) querying $1$-simplex faces of $2$-simplices; 5) querying vertices of simplices and calculating their coordinates in $\Re^n$. The types of simplices we need are, 1) $n$-simplex; 2) $1$-simplex; 3) $2$-simplex. 
The fixed dimensionality of a given $n$ makes it possible to perform these queries on GPU without dynamic memory allocation. 

\usetikzlibrary{shapes.geometric}
\usetikzlibrary{shapes.arrows}
\begin{figure}
    \centering
\begin{tikzpicture}
    [node_style/.style={draw, rectangle},
    data_node/.style={node_style, fill=orange!50, rounded rectangle},
    kernel_node/.style={node_style, fill=blue!40, rounded rectangle},
    arrow_style/.style={thick, -Stealth}]
    \node[node_style,fill=blue!10, dashed, minimum width = 8.8cm, minimum height=5.2cm] (le) at (-0.1cm, -0.5cm) {};
    \node[above left =-0.8cm and -2.2cm of le]{Locate Edges};
    
    \node[above=-0.2cm of le] (marker) {};
    \node [trapezium, trapezium angle=60, minimum width=45mm, draw, thick, above=-1.29cm of marker] {Parallelized for each seed};
        \node[data_node, above=-0.8cm of marker, minimum width=4.5cm, ] (ns){Malloc: $n$-simplices};
        \node[kernel_node, 
        above =-1.83cm of marker, minimum width=5cm] (fn) {Kernel: find all $n$-simplices};

    \node[above=-2.3cm of le] (marker1) {};
    \node [trapezium, trapezium angle=30, minimum width=55mm, draw, thick, above=-1.30cm of marker1] {Parallelized for each $n$-simplex};
        \node[data_node, above=-0.8cm of marker1, minimum width=5.5cm] (se){Malloc: $n$-simplices' edges};
        \node[kernel_node, 
        above =-1.83cm of marker1, minimum width=5cm] (ie) {Kernel: find all interesting $n$-simplices's edges};
        \node[kernel_node, 
        above =-2.9cm of marker1, minimum width=5cm] (he) {Kernel: hash all edges};
        \node[single arrow, draw=black, fill=yellow!50, 
      minimum width = 8pt, single arrow head extend=3pt,
      minimum height=5mm, above=0.28cm of he, rotate=-90] (cpu2gpu) {};
      \node[single arrow, draw=black, fill=yellow!50, 
      minimum width = 8pt, single arrow head extend=3pt,
      minimum height=5mm, above=0.28cm of se, rotate=-90] (cpu2gpu) {};
\end{tikzpicture}
    \caption{The steps for locating edges from seed points.}
    \label{fig:tri}
\end{figure}
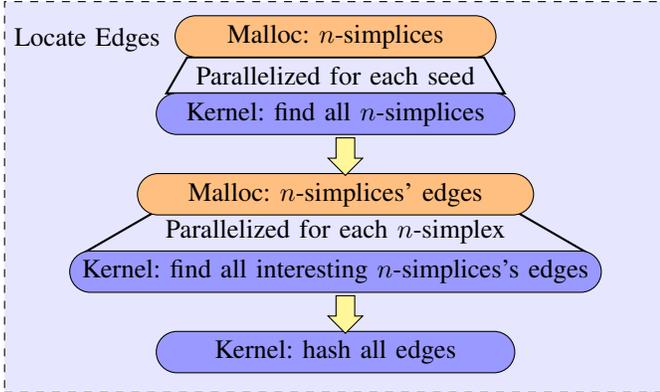

With these GPU versions of querying functions, we can perform the manifold tracing algorithm. We discuss the algorithm in two steps. In the first step, we locate from $\Qseeds$ the edges ($1$-simplices) that intersect the manifold. This step has three GPU kernels. 
In the first kernel, we parallelize for each point in $\Qseeds$ to locate its $n$-simplex. 
We pre-allocate memory for the $n$-simplices since each seed point is contained in one and only one $n$-simplex. 
In the second kernel, we parallelize for each $n$-simplex, query its edges, and save the edges that intersect with the manifold. We pre-allocate memory for all edges. A $n$-simplex has $n(n-1)/2$ edges. 
An edge with vertices $v_1$ and $v_2$ intersects a manifold if $\infproof(v_1) \infproof(v_2) < 0$. Lastly, we save the edges to a hashset to make sure there are no duplicates. An overview of these steps is shown in \autoref{fig:tri} on the left. 

In the second step, we perform graph traversal starting with the edges we saved in the first step. Breadth-first search graph traversal on GPU has been studied before~\cite{fu2014parallel, merrill2015high, liu2016ibfs}. 
A common approach is to expand the current frontiers to find new nodes, remove any duplicates, and add the remaining new nodes as the next level's frontier. We take a similar approach. 
To fully utilize the cores and create less thread divergence, we design the algorithm to perform one task in one kernel. 

In each iteration, we start by parallelizing for each frontier edge and query their cofaces. 
We can calculate the maximum number of cofaces from the permutahedral representation's coface querying process and the fact that our co-dimension is always $1$. 
In $\Re^n$, an edge with partition $\{p_1, p_2\}$ has $(|osp(p+1)| + |osp(p_2)|)$ $2$-simplex cofaces, where $|op(p_i)|$ is the number of ordered two set partitions for $p_i$, which results naturally from the coface querying process of permutahedral representation. 

Next, we parallelize for each $2$-simplex coface and save edges of the cofaces if the edges intersect with the manifold and have not been visited before. We can pre-allocate memory space for the cofaces' edges from the following. Each $2$-simplex coface has three edges, and two and only two edges in a $2$-simplex intersect with the manifold assuming the manifold locally does not pass through a single edge twice. One of the two intersecting edges must have been visited before, which is the edge that we query to find the coface previously. 
This means the total number of new edges equals the number of cofaces. 
Lastly, we hash the new edges to remove any duplication and add the hashed edges as the next level's frontier. An overview of this step is shown in \autoref{fig:search} on the right. 

\usetikzlibrary{shapes.geometric}
\usetikzlibrary{shapes.arrows}
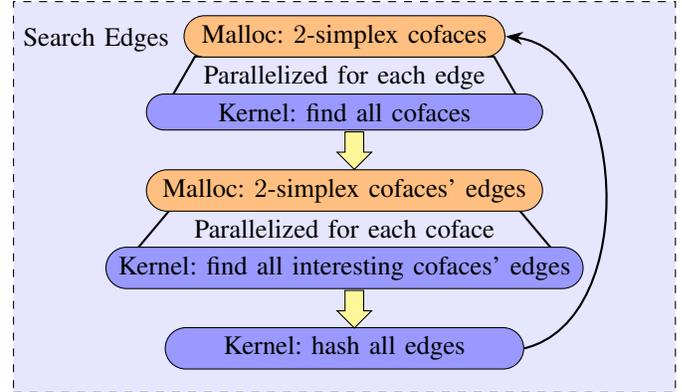
\begin{figure}
    \centering
\begin{tikzpicture}
    [node_style/.style={draw, rectangle},
    data_node/.style={node_style, fill=orange!50, rounded rectangle},
    kernel_node/.style={node_style, fill=blue!40, rounded rectangle},
    arrow_style/.style={thick, -Stealth}]
    \node[node_style,fill=blue!10, dashed, minimum width = 8.8cm, minimum height=5.2cm] (le) at (-0.1cm, -0.5cm) {};
    \node[above left =-0.8cm and -2.2cm of le]{Search Edges};
    
    \node[above=-0.2cm of le] (marker) {};
    \node [trapezium, trapezium angle=60, minimum width=45mm, draw, thick, above=-1.35cm of marker] {Parallelized for each edge};
        \node[data_node, above=-0.8cm of marker, minimum width=4.5cm, ] (ns){Malloc: $2$-simplex cofaces};
        \node[kernel_node, 
        above =-1.77cm of marker, minimum width=5.5cm] (fn) {Kernel: find all cofaces};

    \node[above=-2.25cm of le] (marker1) {};
    \node [trapezium, trapezium angle=50, minimum width=55mm, draw, thick, above=-1.30cm of marker1] {Parallelized for each coface};
        \node[data_node, above=-0.8cm of marker1, minimum width=5.5cm] (se){Malloc: $2$-simplex cofaces' edges};
        \node[kernel_node, 
        above =-1.83cm of marker1, minimum width=5cm] (ie) {Kernel: find all interesting cofaces' edges};
        \node[kernel_node, 
        above =-2.9cm of marker1, minimum width=5cm] (he) {Kernel: hash all edges};
        \node[single arrow, draw=black, fill=yellow!50, 
      minimum width = 8pt, single arrow head extend=3pt,
      minimum height=5mm, above=0.30cm of he, rotate=-90] (cpu2gpu) {};
      \node[single arrow, draw=black, fill=yellow!50, 
      minimum width = 8pt, single arrow head extend=3pt,
      minimum height=5mm, above=0.31cm of se, rotate=-90] (cpu2gpu) {};
      \draw[arrow_style](he.east) to[bend right=80]  (ns.east);

\end{tikzpicture}

    \caption{The steps for searching edges in iteration.}
    \label{fig:search}
\end{figure}

To summarize, because our co-dimension is always $1$, we can eliminate dynamic memory management in our GPU triangulation algorithm design. All memory allocations for new data structures have fixed maximum size. Also, we design the kernels such that all threads run one task at one time, which eliminates as many diverging branches as possible.

\begin{figure}
    \centering
    \begin{tikzpicture}[line join = round, line cap = round]
\coordinate [label={below left:$v_0$}] (v0) at (0, 0);
\coordinate [label={below right:$v_1$}] (v1) at (\pythagwidth, 0);
\coordinate [label={}] (vd) at (\pythagwidth/2, 0);
\coordinate [label={above right:$v_2$}] (v2) at (\pythagwidth, \pythagheight);
\coordinate [label={above right:}] (v01) at (\pythagwidth/2, 0);
\coordinate [label={above right:}] (v02) at (\pythagwidth/2, \pythagheight/2);
\coordinate [label={above right:}] (v12) at (\pythagwidth, \pythagheight/2);
\coordinate [label={above left:$v: [0, 0]$}] (s1) at (-0.1cm, 1.4cm);
\coordinate [label={above left:$p:\{\{0\}, \{1, 2\}\}$}] (s1) at (1.1cm, 1cm);

\path[draw, fill=blue!20!white, line width=1pt, postaction={on each segment={left arrow=black}}]  (v0) -- (v1)--(v2)--cycle;  
\draw[dashed] (v01)--(v02);
\draw[dashed] (v12)--(v02);
\draw[dashed] (v12)--(v01);
\draw[thick,red,-, line width=2pt]  (v0) to (vd);
\end{tikzpicture}
\begin{tikzpicture}[line join = round, line cap = round]
\pgfmathsetmacro{\factor}{0.6};
\coordinate [label=below:$v_0$] (v0) at (0,0,0*\factor);
\coordinate [label=below:$v_1$] (v1) at (2,0,0*\factor);
\coordinate [label=right:$v_2$] (v2) at (2,0,-2*\factor);
\coordinate [label=left:$v_3$] (v3) at (2,2,-2*\factor);

\coordinate [label=below:] (v01) at (1,0,0*\factor);
\coordinate [label=below:] (v02) at (1,0,-1*\factor);
\coordinate [label=below:] (v03) at (1,1,-1*\factor);
\coordinate [label=below:] (v12) at (2,0,-1*\factor);
\coordinate [label=below:] (v13) at (2,1,-1*\factor);
\coordinate [label=below:] (v23) at (2,1,-2*\factor);
\coordinate [label={above left:$v: [0, 0, 0]$}] (s1) at (1.4cm, 1.6cm);

\coordinate [label={above left:$p:\{\{0\}, \{1, 2, 3\}\}$}] (s1) at (1.4cm, 1.2cm);
\draw[dashed] (v01)--(v02);
\draw[dashed] (v12)--(v02);
\draw[dashed] (v12)--(v01);

\draw[dashed] (v01)--(v13);
\draw[dashed] (v13)--(v03);
\draw[dashed] (v03)--(v01);

\draw[dashed] (v23)--(v12);
\draw[dashed] (v12)--(v13);
\draw[dashed] (v13)--(v23);

\draw[dashed] (v03)--(v23);
\draw[dashed] (v02)--(v03);
\draw[dashed] (v23)--(v02);

\draw[-, fill=red!30, opacity=.5] (v0)--(v1)--(v2)--cycle;
\draw[-, fill=blue!30, opacity=.5] (v0) --(v1)--(v3)--cycle;
\draw[-, fill=purple!30, opacity=.5] (v1)--(v2)--(v3)--cycle;
\path[draw, line width=1pt, postaction={on each segment={left arrow=black}}]  (v0) -- (v1)--(v2)--(v3)--cycle; 
\draw[thick,red,-, line width=2pt]  (v0) to (v01);

\end{tikzpicture}
    \caption{First edge (marked in red) and its permutahedral presentation in the new triangulation, for subdivision of a 2-simplex and a 3-simplex, with $k=2$. }
    \label{fig:decompose}

\end{figure}
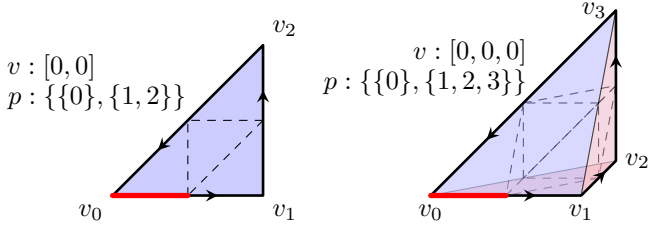

\subsection{Batch Triangulation}
\label{sec:batchtri}
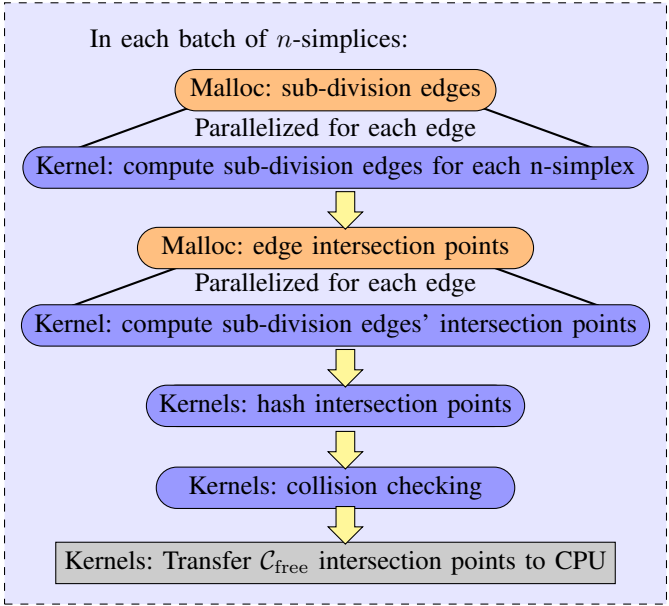
\begin{figure}
    \centering
    \begin{tikzpicture}
    [node_style/.style={draw, rectangle},
    data_node/.style={node_style, fill=orange!50, rounded rectangle},
    kernel_node/.style={node_style, fill=blue!40, rounded rectangle},
    arrow_style/.style={thick, -Stealth}]
    \node[node_style,fill=blue!10, dashed, minimum width = 8.8cm, minimum height=8.0cm] (le) at (-0.1cm, -0.5cm) {};
    \node[above left =-0.8cm and -5.5cm of le]{In each batch of $n$-simplices:};
    
    \node[above=-0.9cm of le] (marker) {};
    \node [trapezium, trapezium angle=20, minimum width=45mm, draw, thick, above=-1.33cm of marker] {Parallelized for each edge};
        \node[data_node, above=-0.8cm of marker, minimum width=4.5cm, ] (ns){Malloc: sub-division edges};
        \node[kernel_node, 
        above =-1.83cm of marker, minimum width=5cm] (fn) {Kernel: compute sub-division edges for each n-simplex};

    \node[above=-3.0cm of le] (marker1) {};
    \node [trapezium, trapezium angle=20, minimum width=55mm, draw, thick, above=-1.30cm of marker1] {Parallelized for each edge};
        \node[data_node, above=-0.8cm of marker1, minimum width=5.5cm] (se){Malloc: edge intersection points};
        \node[kernel_node, 
        above =-1.83cm of marker1, minimum width=5cm] (ie) {Kernel: compute sub-division edges’ intersection points};
        \node[kernel_node, 
        above =-2.9cm of marker1, minimum width=5cm] (he) {Kernel: hash all edges};
        \node[kernel_node, 
        above =-2.9cm of marker1, minimum width=5cm] (he) {Kernels: hash intersection points};
        \node[kernel_node, 
        above =-3.99cm of marker1, minimum width=5cm] (cc) {Kernels: collision checking};
        \node[draw, rectangle, fill=black!20,
        above =-4.99cm of marker1, minimum width=5cm] (tra) {Kernels: Transfer $\CFree$ intersection points to CPU};
        \node[single arrow, draw=black, fill=yellow!50, 
      minimum width = 8pt, single arrow head extend=3pt,
      minimum height=5mm, above=0.28cm of he, rotate=-90] (cpu2gpu) {};
      \node[single arrow, draw=black, fill=yellow!50, 
      minimum width = 8pt, single arrow head extend=3pt,
      minimum height=5mm, above=0.28cm of cc, rotate=-90] (cpu2gpu) {};
      \node[single arrow, draw=black, fill=yellow!50, 
      minimum width = 8pt, single arrow head extend=3pt,
      minimum height=5mm, above=0.28cm of se, rotate=-90] (cpu2gpu) {};
      \node[single arrow, draw=black, fill=yellow!50, 
      minimum width = 8pt, single arrow head extend=3pt,
      minimum height=5mm, above=0.28cm of tra, rotate=-90] (cpu2gpu) {};
\end{tikzpicture}
    \caption{The steps in each batch triangulation of $n$-simplices.}
    \label{fig:batch}
\end{figure}
In the manifold tracing algorithm, with a given manifold in $\Re^n$, the number of intersecting edges ($1$-simplices) is related to the resolution of the $\Re^n$ Coxeter triangulation and the dimension $n$. If we use a smaller resolution, then the algorithm will produce more intersecting edges. If we run in high dimensions, the algorithm also produces more intersecting edges. Because of this, the algorithm sometimes requires more memory to store the data structures (the edges and cofaces in particular) than what the GPU can offer. We resolve this issue with limited memory resources through batch triangulation.
While we cannot change the dimensionality of the problem, we can change the resolution of the $\Re^n$ triangulation, which would produce a coarser manifold triangulation. Then, we perform a finer sub-triangulation on top of the existing triangulation to get to the refinement we need. 

The first step is to find the $n$-simplices that intersect with the manifold from the intersection edges of the triangulation algorithm in \autoref{sec:trigpu}. 
We begin by querying the $n$-simplex cofaces of the intersection edges. 
We can compute the maximum number of $n$-simplex cofaces using the permutahedral presentation. 
In $\Re^n$, an edge with partition $\{p_1, p_2\}$ has $|p1|!|p2|!$ $n$-simplex cofaces, where $|p_i|!$ is the factorial of $p_i$'s size.
Then, we hash the $n$-simplices to remove any duplicates. 
The $n$-simplices that intersect with the manifold decompose the manifold into coarse pieces. Next, we perform finer triangulation within each piece.

The core idea is to decompose each $n$-simplex with symmetric subdivision~\cite{moore1992simplicial}, find all sub-simplices' edges, and then iterate through all the edges to find the ones intersecting with the manifold. To make sure the resulting sub-simplices' edges inside a $n$-simple are the edges of a Coxeter triangulation with finer resolution, we define and find all sub-simplices' edges through permutahedral presentation inside a $n$-simplex.

Assume we want the resolution of the triangulation to be $\lambda_T$, and the coarse triangulation uses a resolution of $\lambda_T * k$, where $k$ is the decompose resolution we need and we decompose each edge of the $n$-simplex into $k$ pieces. In $\Re^n$, given a $n$-simplex in a coarse triangulation, with vertex $[v_0, v_1, ..., v_{n-1}]$ and partition $\{\{p_1\}, \{p_2\} ..., \{p_{n+1}\}\}$, $p_1, p_2, ..., p_{n+1} \in \{1, 2, ..., n+1\}$, we start a new ``local'' triangulation in the simplex. In the local triangulation, the vertex $[v_0, v_1, ..., v_{n-1}]$ becomes $[0, 0, ...,0]$, and we can define our first edge in the new triangulation as 
vertex: $[0, 0, ...,0]$, partition $\{\{0\}, \{1, 2, ..., n\}\}$, which corresponds to the first decomposed line segment on the root system cycle's first edge based on the partition of $n$-simplex. This local edge's representation is the same within all $n$-simplices. \autoref{fig:decompose} shows the subdivision and the first edge in a $2$-simplex and a $3$-simplex, with $k=2$. From this first edge, we can use the same queries we used in the manifold tracing algorithm to find all sub-simplices' edges by first querying the cofaces and then querying the edges.

We also need to ensure the edges we find are in the $n$-simplex. For this, we check that the two vertices of the edge are in the $n$-simplex. Given a vertex $v' = \{v'_0, v'_1, ..., v'_{n-1}\}$ in the local triangulation, The value of $v'_i$ denotes the proportion of the edge $(v_{i+1}, v_i)$ occupied by the vertex.  We can compute the barycentric coordinates of the vertex in the $n$-simplex from the coordinate of the vertex in the original triangulation frame,
\begin{equation}
    \begin{aligned}
    v' &= \Vec{v}_0 + v'_0/k(\Vec{v}_1 - \Vec{v}_0) + v'_1/k(\Vec{v}_2 - \Vec{v}_1) + ... \\
    &+ v'_{n-1}/k(\Vec{v}_{n} - \Vec{v}_{n-1})\\
    &=  (1 - v'_0/k) \Vec{v}_0 + (v'_0/k - v'_1/k) \Vec{v}_1 
    + (v'_1/k - v'_2/k) \Vec{v}_2 \\
    &+ ... + (v'_{n-2}/k - v'_{n-1}/k) \Vec{v}_{n-1} + (v'_{n-1}/k)\Vec{v}_{n},
\end{aligned}
\label{eq:bc}
\end{equation}
which gives the barycentric coordinates $(1 - v'_0/k, v'_0/k - v'_1/k, ..., v'_{n-2} - v'_{n-1}, v'_{n-1}/k)$.

A point is inside a simplex if its barycentric coordinates are all no less than $0$, which gives us, 
\begin{equation}
    \left\{\begin{aligned} 
&1 - v'_0/k > 0,\\
    &v'_0/k - v'_1/k > 0, \\
    &...\\
    &v'_{n-2}/k - v'_{n-1}/k > 0, \\
    &v'_{n-1}/k > 0\\
    \end{aligned}\right. \Rightarrow 
    \left\{\begin{aligned}
&k - v'_0 > 0,\\
    &v'_0 - v'_1 > 0, \\
    &...\\
    &v'_{n-2} - v'_{n-1} > 0, \\
    &v'_{n-1} > 0\\
    \end{aligned}\right.
    \label{eq:local}
\end{equation}
We also multiply both sides by $k$ to remove floating-point computation. With \autoref{eq:local}, we can easily check whether a vertex in the local triangulation is in the $n$-simplex or not. 

The resulting cofaces and edges' permutahedral presentation within a $n$-simplex are the same across all $n$-simplice, so we only need to compute this once and we can use it to sub-divide all the $n$-simplices we computed from the triangulation step. Also, by using permutahedral presentation in the local triangulation, the resulting sub-simplices' edges are the same as the simplices' edges in a triangulation with $\lambda_T$, and we have reached the original triangulation resolution.

With this sub-division, we perform triangulation by batch. We estimate the overall memory usage for the sub-division of each $n$-simplex, and then we compute how many $n$-simplices the current remaining memory of the GPU could work with at one time to put them into batches. Within each batch, we use the sub-division to get the edges' vertices, then compute the intersection points of the edges with the manifold. These kernels are parallelized on each edge. Then, we hash the intersection points to remove any duplicates and check for collision with the remaining intersection points. If there are intersection points in $\CFree$, we transfer those points back to the CPU side to retrain the manifold. We free all GPU memory for the computation related to the batch triangulation and collision checking at the end of each batch, so we can take another batch of $n$-simplices. \autoref{fig:batch}
shows these steps in each batch.

\subsection{Collision checking on GPU}
In \autoref{fig:batch}, we observe that the collision checking is performed by batch on a large set of configurations (the intersection points). This is ideal for collision checking on GPU, where it takes a large number of configurations at one time and parallelizes collision checking of each configuration in each thread. We implemented a forward kinematics (FK) kernel and a collision-checking kernel for batch collision checking on GPU.

The collision-checking kernel is based on primary shapes, more specifically, spheres vs $\{$cylinders, boxes, spheres$\}$. With a new robot and a new scene, we need to configure the robot's geometries into enclosing spheres and represent the world geometries using cylinders, boxes, and/or spheres. 
Our geometry checking is based on exact shapes instead of bounding boxes~\cite{pan2012gpu, sundaralingam2023curobo}, see Appendix \ref{app:cc} for detail. 
The threads are first parallelized for each sphere forming the robot, then for each configuration. When there are too many points to check such that there is not enough memory to hold all the FK and intermediate collision checking results, we perform collision checking by batch. 

\subsection{Analysis}
\label{sec:analysis}
Our computational design minimizes data transfer between the CPU and GPU. For triangulation, we complete the triangulation part entirely on GPU and only require the data for the manifold we are triangulating to be transferred to GPU. For batch collision checking after triangulation, we only need geometry information about the scene and the robot. The time required for data transfer would not increase with dimensionality. In practice, memory usage is less than $5\%$ of total GPU time. 

We emphasize that most of the kernels we use for triangulation and collision checking have very high GPU occupancy. The kernels for finding $n$-simplice, finding new edges of $n$-simplices, finding cofaces of edges, and hashing new edges have $100\%$ theoretical occupancy. Finding new edges of cofaces does not have $100\%$ theoretical occupancy because we also check for whether the edge intersects with the manifold. Calculating intersection points of edges also does not have $100\%$ theoretical occupancy because not all threads have the same number of iterations for the computation. 
Because of the high GPU occupancy, the algorithm demonstrates strong scalability with increased GPU cores and memories. With more powerful GPUs, infeasibility proofs for 7-DoF robots can be completed within a reasonable time. 

\begin{figure*}
 \centering
 \begin{tabular}{@{}c@{}}
{\frame{\begin{overpic}[width=0.3\linewidth, clip=true, trim = 184mm 110mm 182mm 90mm]{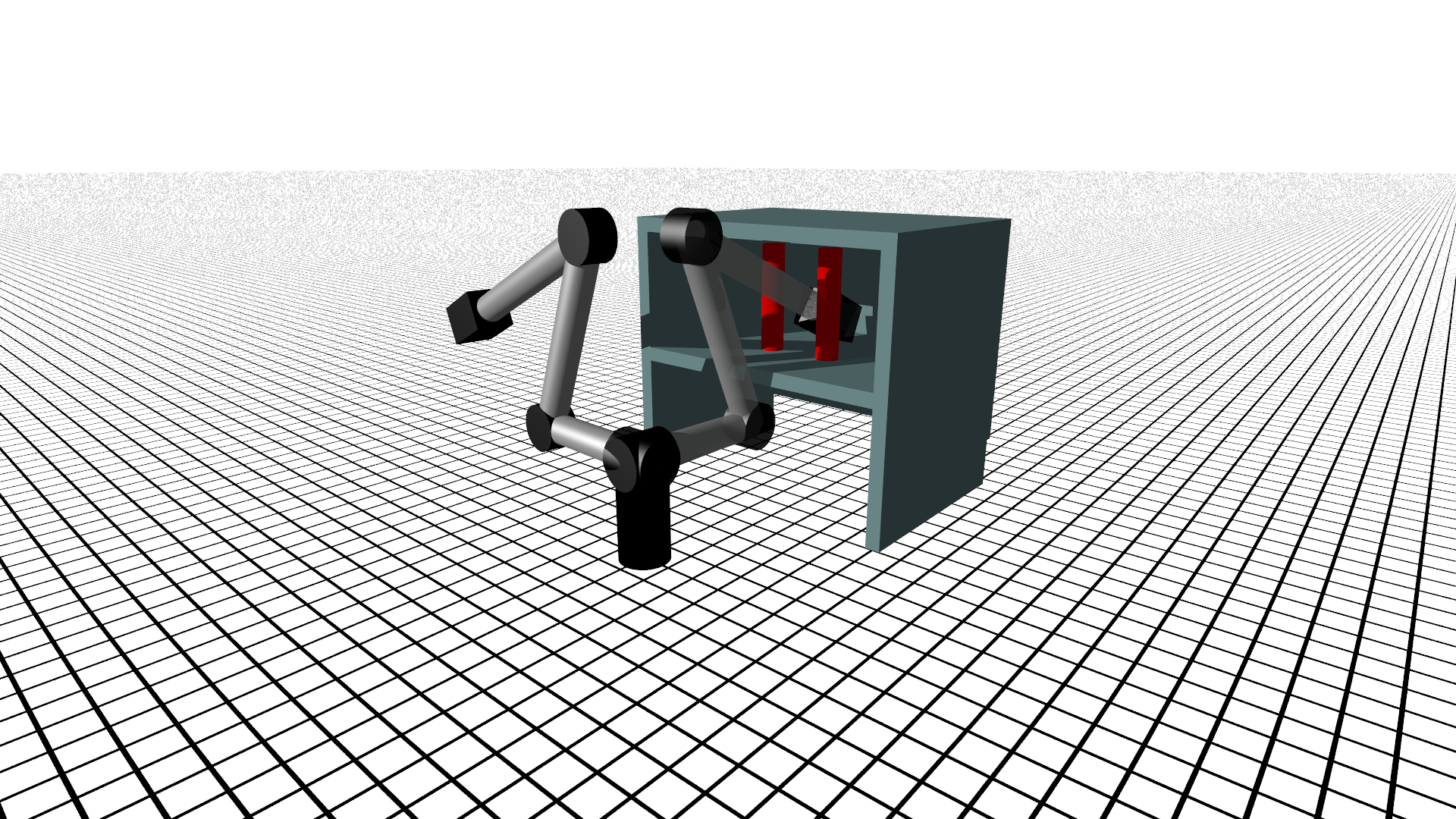}
        \put(5,27){\colorbox{white}{%
            \parbox{0.05\linewidth}{ \color{black}START} }}
            \put(62,12){\colorbox{white}{%
            \parbox{0.05\linewidth}{ \color{black}GOAL} }}
        \end{overpic}}} \\[\abovecaptionskip]
    \small (a) PackBot Scene
          \label{fig:packbot}
  \end{tabular}
 \begin{tabular}{@{}c@{}}
{\frame{\begin{overpic}[width=0.3\linewidth, clip=true, trim = 54mm 57mm 72mm 10mm]{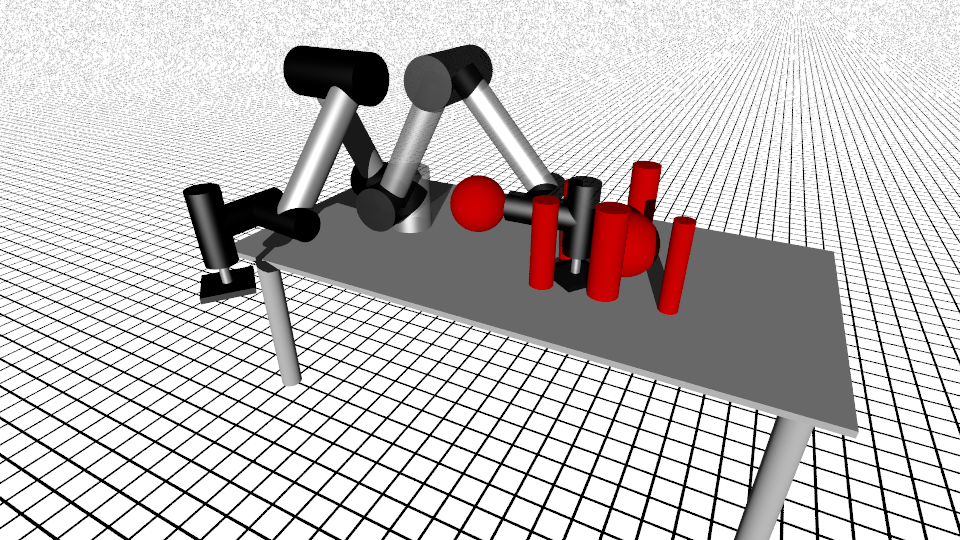}
        \put(4,7){\colorbox{white}{%
            \parbox{0.06\linewidth}{ \color{black}START} }}
            \put(65,45){\colorbox{white}{%
            \parbox{0.05\linewidth}{ \color{black}GOAL} }}
        \end{overpic}}} \\[\abovecaptionskip]
    \small b) Universal Robot Scene
          \label{fig:packbot}
  \end{tabular}
 \begin{tabular}{@{}c@{}}
{\frame{\begin{overpic}[width=0.3\linewidth, clip=true, trim = 24mm 20mm 42mm 12mm]{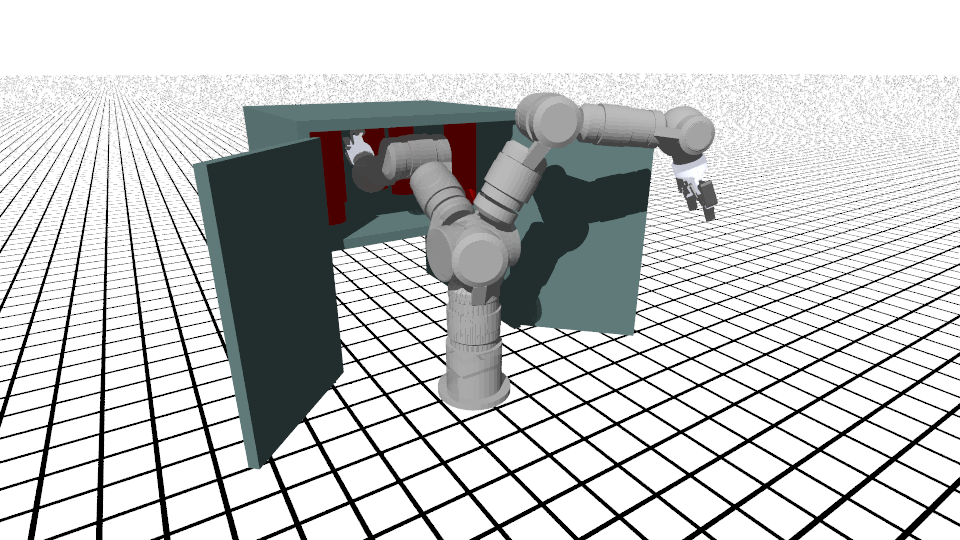}
        \put(4,37){\colorbox{white}{%
            \parbox{0.05\linewidth}{ \color{black}GOAL} }}
            \put(68,22){\colorbox{white}{%
            \parbox{0.05\linewidth}{ \color{black}START} }}
        \end{overpic}}} \\[\abovecaptionskip]
    \small c) Schunk Robot Scene
          \label{fig:packbot}
  \end{tabular}
      \caption{Experiment Scenes. The packbot scene is 5-DoF. The universal robot scene run in 5-D0F and 6-DoF. The Schunk scene is 6-DoF. We experiment with 4 scenes in total.}
\label{fig:exp}
\end{figure*}
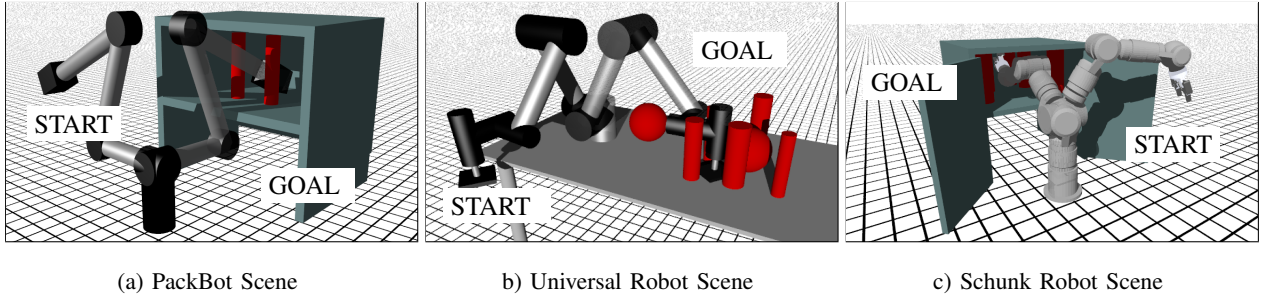

\section{Experiment}
\label{sec:exp}
In this section, we validate our algorithm through experiments. 
We run two 5-DoF scenes, similar to the configuration outlined in our prior publication~\cite{li2023scaling}, to evaluate the speedup achieved through GPU implementation. We also run two 6-DoF scenes to show the scalability of our algorithm. 

We run our
experiment on a desktop system with an NVIDIA GeForce RTX 4070 GPU with 12G memory and an Intel i9-13900K CPU with 48 cores.  We use
PRM~\cite{kavraki1996probabilistic} in OMPL~\cite{sucan2012open} as the sampling-based motion planner that runs in parallel with our infeasibility proof construction, and get samples from the planner to train the manifold and sample the manifold. We solve the nonlinear optimization problems using
sequential least-squares quadratic programming
(SLSQP)~\cite{kraft1988software,kraft1994algorithm} in
NLopt~\cite{nlopt}.  We adopt the Coxeter triangulation module in
GUDHI~\cite{gudhi:urm}, and implemented our own GPU kernels for permutahedral representation with necessary queries for triangulation. Our triangulation, batch triangulation, and collision checking are all based on CUDA~\cite{nickolls2008scalable}.  We train the RBF-kernel SVM
using ThunderSVM~\cite{wenthundersvm18}, which supports
GPU-accelerated SVM training.  We model robot kinematics using
Amino~\cite{dantam2020robust}, which is interpreted on the CPU, and then geometry information is transferred to the GPU.

\subsection{5-DoF experiments}
We have two 5-DoF experiment scenes, using a similar setup as described in our previous paper~\cite{li2023scaling}.  We configure the robot with inscribing spheres for fast collision checking on GPU. The scenes and the start and goal configurations are shown in \autoref{fig:exp}. The first scene is with Packbot~\cite{packbot510}. The second experiment uses the Universal robot~\cite{universal} with the last joint fixed to make it 5-DoF.  

\subsection{6-DoF experiments}
We have two 6-Dof experiment scenes, as shown in \autoref{fig:exp}. The first scene uses the same universal robot scene with all 6 joints. The second scene uses a Schunk LWA4D manipulator, with the third joint fixed to make it 6-DoF. The scene is shown in \autoref{fig:exp}. The goal in the scene is to reach inside a shelf from a location of the arm outside of the shelf. 

\begin{table}
    \centering
\begin{tabular}{|c|
c
V{3} c |c |}
 \hline
  & \tabtm{Total (s)} & \tablbl{Tri (s)} & \tablbl{Check (s)}\\
  \hline

\tabtm{Packbot} & 22.16 $\pm$14.88 & 5.56 $\pm$6.03 & 1.62 $\pm$2.13\\
\hline


  \hline
\tabtm{Universal-5} & 8.42 $\pm$2.30 & 0.40 $\pm$0.17 & 0.32 $\pm$0.19\\
\hline


  \hline
\tabtm{Universal-6} & 58.62 $\pm$29.66 & 2.86 $\pm$1.71 & 11.80 $\pm$6.29\\
\hline


  \hline
\tabtm{Schunk} &58.12 $\pm$37.75 & 12.65 $\pm$7.97 & 34.49 $\pm$28.33\\
\hline
\end{tabular}
\caption{Runtime results for 5-DOF and 6-DOF manipulators, mean (s) $\pm$STD, averaged over 30 trials. ``Tri" is for triangulation, including batch triangulation. ``Check" is for collision checking. }
    \label{tab:exp}
\end{table}
We summarize the experimental results in \autoref{tab:exp}. For the two 5-DoF scenes, compared with previous results, we achieve more than two orders of magnitude speed up for collision checking in both scenes. The triangulation time in the 5-DoF universal robot scene is also two orders of magnitude faster. The triangulation in the Packbot scene takes slightly longer because the ``size'' of the manifold is larger with a wider goal region in the configuration space. 
The two 6-DoF scenes both take less than 1 minute on average, triangulation takes several seconds to tens of seconds, and collision checking takes tens of seconds.

\subsection{Discussion}
Comparing the 5-DOF Packbot scene and the 6-DoF Universal robot scene, we can see that the triangulation for the Packbot takes longer than the triangulation for the 6-DoF Universal robot. This is caused by the size of the manifold. A larger manifold takes much longer to triangulate than a smaller manifold in the same dimension. The reason for this is that the triangulation algorithm runtime is sensitive to the output size~\cite{kachanovich2019meshing} since we need to trace the entire manifold to find the triangulation. This also becomes more noticeable in higher dimensions, as the manifold also becomes higher dimension. 

We also noticed some delays on the planner side. The results for the Packbot, and the Universal robot in 5-DoF and 6-DoF show a difference between the sum of the triangulation and checking time and the total planning time. The primary reason for this difference is that after checking, we provide all the $\CFree$ manifold points back to the planner side, and the planner needs to process these $\CFree$ points and add them to the planning graph. This step helps with the convergence of the manifold, which is the basis of a successful triangulation and checking. If the triangulation and checking on GPU find a larger number of points, then the planner would take a longer time to process those points, thus causing the delay. 


As we mentioned in \autoref{sec:analysis}, many kernels in our algorithm can achieve very high theoretical occupancy and achieved occupancy. In practice, many factors can effect GPU occupancy, but we still achieve very high measured GPU occupancy.
Finding new edges of n-
simplices achieves an occupancy of around 80\%; 
finding cofaces of edges achieves an occupancy of around 88\%;
finding new edges of cofaces achieves an occupancy of around 80\%;
and calculating
intersection points of edges achieves an occupancy of around 53\%. Calculating intersection points has lower occupancy because the calculation is based on an iterative algorithm that terminates when a desired accuracy is reached, which could have a lot of branching. 

\section{Future Work and Conclusion}




\bibliographystyle{IEEEtran}
\bibliography{IEEEabrv, references, dyalab-pubs, dyalab-refs}

\appendices

\section{Collision Checking}
\label{app:cc}
In this appendix, we record the exact collision checking method between sphere and primacy shapes (box, cylinder, sphere) for our batch collision checking. Sphere vs sphere can be easily checked, so we omit this case. 

\subsection{Sphere vs Box}
The box has dimension $[l_x, l_y, l_z]$, and coordinate frame $\{F_b\}$. The origin of this frame is at the geometric center of the box and the axes of the frame align with the three edges of the box. 
The sphere is located at $\Vec{c_s}$, with a radius of $r$.  
We first transform the sphere into the box's coordinate frame, which gives the center of the sphere at $^b\Vec{c_s}$ in $\{F_b\}$. 
Because the box is symmetric for all three axes, we can map the center of the sphere to the first octant by computing the absolute values of x, y, z in $^b\Vec{c_s}$, we note this mapped coordinate with $\{p_{x}, p_{y}, p_{z}\}$. We only need to consider the first octant from now on. 

Exact collision checking between sphere and box three steps.
In the first step, we check whether 
\begin{equation}
    \exists i \in \{x, y, z\}, \quad \mathrm{s.t.} \quad p_i - l_i/2 > r,
    \label{eq:sb-bb}
\end{equation}
which means the sphere center is located outside a bounding box region of radius plus the box dimensions, so the sphere does not collide with the box. We have $l_i/2$ since we are only considering the box's part in the first octant at this point. 

\begin{figure}
\centering
    \begin{tikzpicture}
\pgfmathsetmacro{\cubex}{2.5}
\pgfmathsetmacro{\cubey}{2}
\pgfmathsetmacro{\cubez}{2}
\pgfmathsetmacro{\axisl}{0.7}
\draw[->, thick] (0,0,0) -- (\cubex + \axisl,0,0) node[right] {$x$};
\draw[->, thick] (0,0,0) -- (0, 0, \cubez + 2*\axisl) node[below] {$z$};
\draw[->, thick] (0,0,0) -- (0, \cubey + \axisl, 0) node[above] {$y$};


\draw[-,fill=blue!30] (0,0,\cubez) -- ++(\cubex,0,0) -- ++(0,\cubey,0) -- ++(-\cubex,0,0) -- cycle;
\draw[-,fill=blue!30] (\cubex,0,\cubez) -- ++(0,0,-\cubez) -- ++(0,\cubey,0) -- ++(0,0,\cubez) -- cycle;
\draw[-,fill=blue!30] (\cubex,\cubey,\cubez) -- ++(-\cubex,0,0) -- ++(0,0,-\cubez) -- ++(\cubex,0,0) -- cycle;
\draw[dashed, fill=yellow!30, opacity=.5] (\cubex,\cubey,\cubez) -- ++(0,0,-\cubez) -- ++(0,\cubey,0) -- ++(0,0,2*\cubez) -- ++(0,-2*\cubey,0) --++(0, 0, -\cubez) -- cycle;

\draw[dashed, fill=yellow!30, opacity=.5] (0,\cubey,\cubez) -- ++(0,0,\cubez) -- ++(\cubex * 2,0,0) -- ++(0,0,-2*\cubez) -- ++(-\cubex,0,0) --++(0, 0, \cubez) -- cycle;

\draw[dashed, fill=yellow!30, opacity=.5] (0,\cubey,\cubez) -- ++(0,\cubey,0) -- ++(\cubex * 2,0,0) -- ++(0,-2*\cubey,0) -- ++(-\cubex,0,0) --++(0, \cubey, 0) -- cycle;

\fill[yellow!30, opacity=.5] (0,\cubey,\cubez) -- ++(0,\cubey,0) -- ++(0,0,\cubez) --++(0, -\cubey, 0) -- cycle;

\fill[yellow!30, opacity=.5] (\cubex,0,\cubez) -- ++(\cubex,0,0) -- ++(0,0,\cubez) --++(-\cubex, 0, 0) -- cycle;

\fill[yellow!30, opacity=.5] (\cubex,\cubey,0) -- ++(\cubex,0,0) -- ++(0,\cubey,0) --++(-\cubex, 0, 0) -- cycle;

\draw[dashed,fill=red!30, opacity=.5] (\cubex,\cubey,\cubez) -- ++(\cubex,0,0) -- ++(0,\cubey,0) -- ++(-\cubex,0,0) -- cycle;
\draw[dashed,fill=red!30, opacity=.5] (\cubex,\cubey,2*\cubez) -- ++(0,0,-\cubez) -- ++(0,\cubey,0) -- ++(0,0,\cubez) -- cycle;
\draw[dashed,fill=red!30, opacity=.5] (2*\cubex,\cubey,2*\cubez) -- ++(-\cubex,0,0) -- ++(0,0,-\cubez) -- ++(\cubex,0,0) -- cycle;


\draw[dashed] (0, 0, 0)--(0, \cubey, 0);
\draw[dashed] (\cubex, 0, 0)--(0, 0, 0);
\draw[dashed] (0, 0, \cubez)--(0,0,0);
\coordinate [label={left:$l_y / 2$}] (ly) at (0, \cubey/2, \cubez);
\coordinate [label={left:$l_z / 2$}] (ly) at (0, \cubey, \cubez/2);
\coordinate [label={below:$l_x / 2$}] (ly) at (\cubex/2-0.2, -0.4, \cubez/2);
\coordinate [label={below:$r$}] (ly) at (\cubex/2*3, \cubey, \cubez);
\coordinate [label={right:$r$}] (ly) at (\cubex, \cubey/2*3, \cubez);
\coordinate [label={right:$r$}] (ly) at (\cubex, \cubey, \cubez*3/2);

\end{tikzpicture}
    \caption{Collision Checking for Sphere vs Box, in the first cotant. }
    \label{fig:sb-cc}
\end{figure}
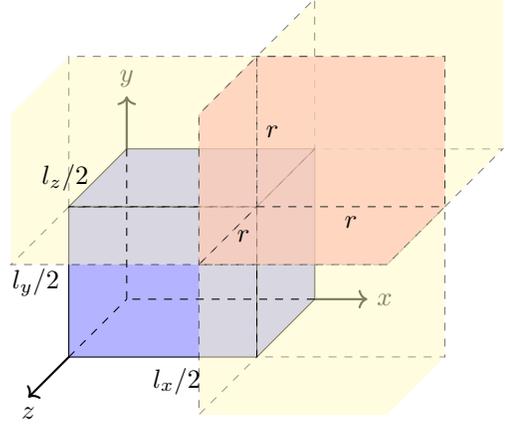

If \autoref{eq:sb-bb} is not true, then we check whether the sphere center is in the ``edge'' region or the ``vertex'' region of the box. 
For a sphere center in the edge regions, we have 
\begin{equation}
\begin{aligned}
    &(p_i < l_i/2) \land (p_j > l_j/2) \land (p_k > l_k/2),\\ &i,j,k \in \{x, y, z\}, i \neq j \neq k.
\end{aligned}
\label{eq:edge-region}
\end{equation}
The edge regions are shown as the yellow regions in \autoref{fig:sb-cc}.  
For sphere centers in edges regions following \autoref{eq:edge-region}, if
\begin{equation}
    (p_j - l_j/2)^2 + (p_k - l_k/2)^2 > r^2,
\end{equation}
which means the distance of the sphere center to the edge is larger than the radius, then the sphere does not collide with the box. 

For a sphere center in the vertex region, we have 
\begin{equation}
p_i > l_i/2, \forall i \in \{x, y, z\}.
\label{eq:vertex-region}
\end{equation}
The vertex region is shown as the red region in \autoref{fig:sb-cc}.  
For sphere centers in the vertex region, if 
\begin{equation}
    \sum_{i \in \{x, y, z\}}(p_i > l_i/2)^2 > r^2, 
\end{equation}
which means the distance of the sphere center to the vertex is larger than the radius, then the sphere does not collide with the box. 
If none of the above-mentioned cases are true, the sphere collides with the box. 

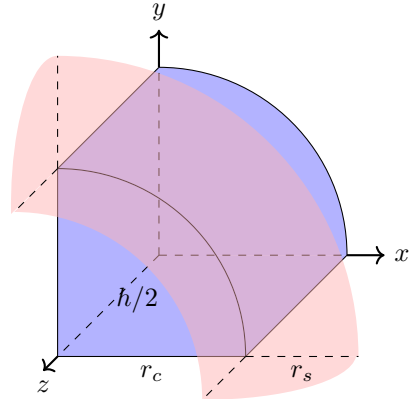
\begin{figure}
    \centering
    	\begin{tikzpicture}
     \def\r{2.5}
     \def\rs{1.5}
           \def\h{3.5}
		\draw[->, thick] (\r,0,0) -- (3,0,0) node[right] {$x$};
\draw[->, thick] (0,0,\h) -- (0,0,4) node[below] {$z$};
\draw[->, thick] (0,\r,0) -- (0,3,0) node[above] {$y$};
           
		\fill[fill=blue!30] (0, 0, 0) -- (\r, 0, 0) arc [x radius = \r, y radius =\r, start angle=0, end angle=90] -- cycle;
  \draw[-] (\r, 0, 0) arc [x radius = \r, y radius =\r, start angle=0, end angle=90];
  \fill[fill=blue!30] (0, 0, 0) --++ (0, \r, 0) --++ (0, 0, \h) --++ (0, -\r, 0) -- cycle;
  \fill[fill=blue!30] (0, 0, 0) --++ (\r, 0, 0) --++ (0, 0, \h) --++ (-\r, 0, 0) -- cycle;
\draw[-,fill=blue!30] (0, 0, \h) -- (\r, 0, \h) arc [x radius = \r, y radius =\r, start angle=0, end angle=90] --cycle;
\draw[-] (\r, 0, \h) --++ (0, 0, -\h);
\draw[-] (0, \r, \h) --++ (0, 0, -\h);
\draw[dashed] (0, 0, 0) --++ (0, 0, \h);
\draw[dashed] (0, 0, 0) --++ (\r, 0, 0);
\draw[dashed] (0, 0, 0) --++ (0, \r, 0);
\fill[fill=red!30, opacity=.5] (\r, 0, \h) -- (\r+\rs, 0, \h) arc [x radius = \r+\rs, y radius =\r+\rs, start angle=0, end angle=90] --++ (0, -\rs, 0) arc [x radius = \r, y radius =-\r, start angle=-90, end angle=0] -- cycle;
\fill[fill=red!30, opacity=.5]  (\r, 0, \rs+\h) arc [x radius = \r, y radius =\r, start angle=0, end angle=90]  -- (0, \r, \h) arc [x radius = \r, y radius =-\r, start angle=-90, end angle=0] --cycle;
\fill[fill=red!30, opacity=.5] (\r,0,\h+\rs) arc [start angle=-90,end angle=-4,x radius=2.05,y radius=0.62]-- (\r, 0, \h) --cycle;
\draw[dashed] (\r, 0, \h) -- (\r+\rs, 0, \h);
\draw[dashed] (\r, 0, \h) --++ (0, 0, \rs);
\fill[fill=red!30, opacity=.5] (0,\r+\rs,\h) arc [start angle=90,end angle=183,x radius=0.62,y radius=2]-- (0, \r, \h) --cycle;
\draw[dashed] (0, \r, \h) --++ (0, \rs, 0);
\draw[dashed] (0, \r, \h) --++ (0, 0, \rs);
\coordinate [label={below:$h / 2$}] (h) at (0, 0, \h/2-1);
\coordinate [label={below:$r_c$}] (h) at (\r/2, 0, \h);
\coordinate [label={below:$r_s$}] (h) at (\rs/2 + \r, 0, \h);

	\end{tikzpicture}
    \caption{Collision Checking for Sphere vs Cylinder, in the first cotant. }
    \label{fig:sc-cc}
\end{figure}

\subsection{Sphere vs Cylinder}
The cylinder has height $h$, and radius $r_c$, and coordinate frame $\{F_c\}$. The origin of this frame is at the geometric center of the cylinder and the z-axis of the frame aligns with the central axis of the cylinder. 
Same as before, we first transform the sphere into the cylinder's coordinate frame, then map the sphere to the first octant, we note this mapped coordinate with $\{p_{x}, p_{y}, p_{z}\}$. We only need to consider the first octant in the following. 

Exact collision checking between sphere and cylinder two steps.
In the first step, we check whether 
\begin{equation}
    p_z - h/2 > r_s \lor \sqrt{p_x^2 + p_y^2} > r_s + r_c
    \label{eq:sc-bb}
\end{equation}
which means the sphere center is located outside a padded cylindrical region of radius plus the cylinder dimensions, so the sphere does not collide with the cylinder. 

If \autoref{eq:sc-bb} is not true, then we check whether the sphere center is in the ``ring'' region of the cylinder. 
For a sphere center in the ring regions, we have 
\begin{equation}
(p_z > h/2) \land (\sqrt{p_x^2 + p_y^2} > r_c)
\label{eq:ring-region}
\end{equation}
The ring region is shown as the red region in \autoref{fig:sc-cc}.  
For sphere centers in ring region following \autoref{eq:ring-region}, if
\begin{equation}
    (\sqrt{(p_x^2 + p_y^2)} - r_c)^2 + (p_z - h/2)^2 > r_s^2,
\end{equation}
which means the distance of the sphere center to the edge of the cylinder is larger than the radius of the sphere, the sphere does not collide with the cylinder. 
If none of the above-mentioned cases are true, the sphere collides with the cylinder.

\end{document}